\pdfoutput=1
\documentclass[11pt]{article}
\usepackage[final]{acl}
\usepackage[T1]{fontenc}

\usepackage[utf8]{inputenc}

\usepackage{microtype}

\usepackage{times}
\usepackage{latexsym}
\usepackage[switch,mathlines]{lineno}

\usepackage{hyperref}
\usepackage{setspace}
\usepackage{float}
\usepackage{amsmath}
\usepackage{amsthm}
\usepackage{amssymb}
\usepackage[ruled, vlined, linesnumbered]{algorithm2e}%
\usepackage{algpseudocode}
\usepackage{booktabs}
\usepackage{adjustbox}
\usepackage{colortbl}
\usepackage{multirow}
\usepackage{todonotes}
\usepackage{booktabs}
\usepackage{microtype}

\newcommand{\Ucal}{\mathcal{U}}
\newcommand{\Vcal}{\mathcal{V}}
\newcommand{\Lcal}{\mathcal{L}}
\newcommand{\Xcal}{\mathcal{X}}
\newcommand{\Ycal}{\mathcal{Y}}
\newcommand{\Scal}{\mathcal{S}}
\newcommand{\bfl}{\mathbf l}
\newcommand{\bfx}{\mathbf x}

\newcommand{\lflabel}{\emph{labeled-set}}

\newcommand{\vallabel}{\emph{validation set}}
\newcommand{\sslabel}{\emph{supervised set}}

\newcommand{\ssalgo}{\emph{Supervised}}

\newcommand{\bilabel}{\emph{validation set}}

\usepackage{multirow}

\newcommand{\ouralgo} {\textsc{Wisdom}}
\newcommand{\snuba} {\textsc{Snuba}}
\newcommand{\snubaLFGen} {\textsc{SnubaLFGen}}

\newcommand{\spear} {\textsc{Spear}}
\newcommand{\cage} {\textsc{Cage}}
\newcommand{\snorkel} {\textsc{Snorkel}}

\title{Learning to Robustly Aggregate Labeling Functions for Semi-supervised Data Programming}
\author{Ayush Maheshwari \textsuperscript{1}\thanks{~~Equal contribution}, Krishnateja Killamsetty \textsuperscript{2}\footnotemark[1], Ganesh Ramakrishnan \textsuperscript{1},
\\ \textbf{Rishabh Iyer\textsuperscript{2}, Marina Danilevsky\textsuperscript{3} and Lucian Popa\textsuperscript{3} } \\
\textsuperscript{1}Indian Institute of Technology Bombay, India \\ \textsuperscript{2} The University of Texas at Dallas  \\
\textsuperscript{3} IBM Research – Almaden \\
\texttt{\{ayusham, ganesh\}@cse.iitb.ac.in}\\
\texttt{\{krishnateja.killamsetty, rishabh.iyer\}@utdallas.edu}\\
\texttt{\{mdanile, lpopa\}@us.ibm.com}\\
}

\begin{document}
\maketitle
\begin{abstract}
A critical bottleneck in supervised machine learning is the need for large amounts of labeled data which is expensive and time-consuming to obtain. Although a small amount of labeled data cannot be used to train a model, it can be used effectively for the generation of human-interpretable \textit{labeling functions} (LFs). These LFs, in turn, have been used to generate a large amount of additional noisy labeled data in a paradigm that is now commonly referred to as data programming. Previous methods of generating LFs do not attempt to use the given labeled data further to train a model, thus missing opportunities for improving performance. Additionally, since the LFs are generated automatically, they are likely to be noisy, and naively aggregating these LFs can lead to suboptimal results. In this work, we propose an LF-based bi-level optimization framework \ouralgo{} to solve these two critical limitations. \ouralgo{} learns a \emph{joint model} on the (same) labeled dataset used for LF induction along with any unlabeled data in a semi-supervised manner, and more critically, reweighs each LF according to its goodness, influencing its contribution to the semi-supervised loss using a robust bi-level optimization algorithm. We show that \ouralgo{} significantly outperforms prior approaches on several text classification datasets. The source code can be found at \url{https://github.com/ayushbits/robust-aggregate-lfs}.
\end{abstract}


\section{Introduction}

Supervised machine learning approaches require large amounts of labeled data to train robust machine learning models. Human-annotated \textit{gold} labels have become increasingly important to modern machine learning systems for tasks such as spam detection, (movie) genre classification, sequence labeling, {\em{etc.}}
The creation of labeled data is, however, a time-consuming and costly process that requires large amounts of human labor. Together with the heavy reliance on labeled data for training models, this serves as a deterrent to achieving comparable performance on new tasks. As a result, various methods such as semi-supervision, distant supervision, and crowdsourcing have been proposed to reduce reliance on human annotation.

In particular, several recent data programming approaches \cite{bach2019snorkel, spear, oishik, awasthi2020learning} have proposed the use of \textit{human-crafted} labeling functions to \textit{weakly} associate labels with the training data. Typically, users encode supervision as rules/guides/heuristics in the form of labeling functions (LFs) that assign noisy labels to the unlabeled data, thus reducing dependence on human-labeled data. 
The noisy labels were aggregated using Label aggregators, which often employ generative models, to assign a label to the data instance. Examples of label aggregators are \snorkel~\cite{ratner} and \cage~\cite{oishik}. These models provide consensus on the noisy and conflicting labels assigned by the discrete LFs to help determine the correct labels probabilistically. We could use the obtained labels to train any supervised model/classifier and evaluate on a test set. Apart from the cascaded approach described above, recently proposed semi-supervised paradigm ~\cite{awasthi2020learning, spear} learns to aggregate labels using both features and a very small labeled set in addition to labeling functions. Such approaches have been shown to outperform the completely unsupervised data programming approaches described above.


Data programming (unsupervised or semisupervised) requires \textit{carefully} curated LFs, generally expressed in the form of regular expressions or conditional statements.
Even though creating LFs can potentially take less time than creating large amounts of supervised data, it requires domain experts to spend considerable time identifying and determining the patterns that should be incorporated into LFs. In this paper, we circumvent the requirement of human-curated LFs by instead automatically generating human-interpretable LFs as compositions of simple propositions on the data set by leveraging \snuba{} ~\cite{snuba} which utilizes a small \lflabel{} to induce LFs automatically.
However, as we will show, \snuba{} suffers from two critical limitations, which keep it from outperforming even a simple supervised baseline that is trained on the same \lflabel{}. First, \snuba{} only uses the \lflabel{} to generate the LFs but does not make effective use of it in the final model training.  Secondly, as it naively aggregates these LFs, it is not able to distinguish between very noisy LFs and more useful ones. This work addresses both of these limitations.
\begin{table}[]
\begin{tabular}{|c|c|c|}
\hline
Label                & Generated LFs & Weighting    \\ \hline
\textbf{\small{ENTITY}}      & what does          & {\color{green}$\uparrow$} \\ \hline
\textbf{\small DESCRIPTION} & what is       & {\color{red}$\downarrow$}   \\ \hline
\textbf{\small NUMERIC}     & how long      & {\color{green}$\uparrow$ }  \\ \hline
\textbf{\small DESCRIPTION} & how           & {\color{red}$\downarrow$} \\ \hline
\textbf{\small HUMAN}       & who           & {\color{green}$\uparrow$}   \\ \hline
\textbf{\small DESCRIPTION} & what kind & {\color{red}$\downarrow$} \\ \hline
\textbf{\small LOCATION} & city & {\color{green}$\uparrow$} \\ \hline
\end{tabular}
\caption{Illustration of induced LFs, including examples of 
the issue 
of conflicting LFs, on the TREC dataset. 
Learning importance (weights) of LFs can be used to reduce the conflicts among LFs.}
\label{tab:compare-lfs}
\end{table}

In Table \ref{tab:compare-lfs}, we present a sample set of induced LFs and assigned labels for the TREC dataset~\cite{trec}. The induced LFs are likely to be less precise compared with those created by humans, and they are likely to have more mutual conflicts. Since the LFs are incomplete and noisy, existing label aggregators that merely consume their outputs do not perform well when dealing with such noisy LFs ({\em c.f.} Table~\ref{tab:compare-lfs}). For instance, the sentence \texttt{How long does a dog sleep ?} will be assigned both \textbf{\small DESCRIPTION} and \textbf{\small NUMERIC} labels due to the LFs \textit{how} and \textit{how long}.

As a solution, \textit{how} should be given less importance due to its noisy and conflicting nature, whereas \textit{how long}, associated with the \textbf{\small NUMERIC} label, should be given higher importance. In this paper, we present a bi-level optimization framework for reweighting the induced LFs, which effectively reduces the weights of noisy labels while simultaneously increasing the weights of the more useful ones.  


\begin{figure}
    \centering
    \includegraphics[width=0.48\textwidth]{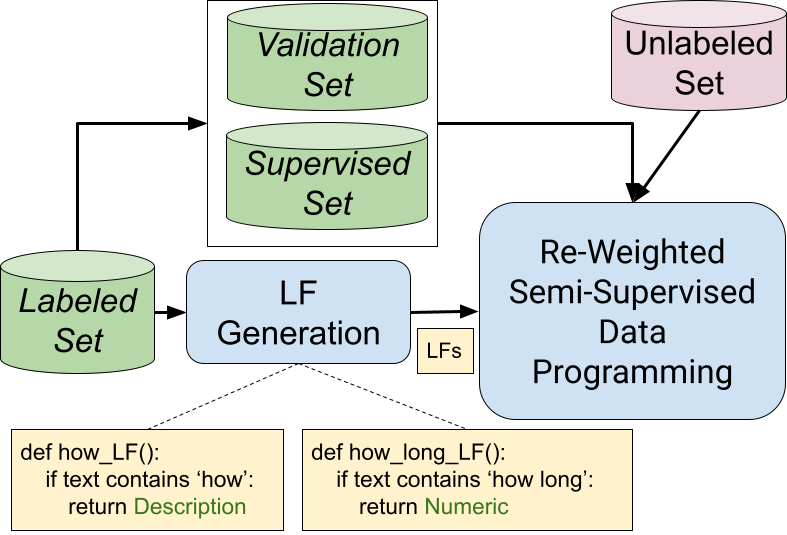}
    \caption{Pictorial depiction of our \ouralgo{} workflow. A small \lflabel{} is used to automatically induce LFs. This labeled set is split equally into \sslabel{} and \vallabel{} to be used by our re-weighted semi-supervised data programming algorithm along with the unlabeled set. }
    \label{fig:arch}
\end{figure}
In Figure~\ref{fig:arch}, we present an overview of our approach. We leverage semi-supervision in the feature space for more effective data programming using the induced (automatically generated) labeling functions.
To enable this, we split the same \lflabel{} (which was used to generate the LFs) into a \sslabel{} and \vallabel{}. The \sslabel{} is used for semi-supervised data programming, and \vallabel{} is used 
to tune (reweight) the LFs.
As a basic framework for semi-supervised data programming, we leverage \spear~\cite{spear}, which has achieved state-of-the-art performance. While the semi-supervised data programming approach helps in using the labeled data more effectively, it  does not solve the problem of noise associated with the LFs. To address this, we propose an LF reweighting framework, \ouralgo{}\footnote{Expanded as re{\bf W}e{\bf I}ghting based {\bf S}emi-supervised {\bf D}ata pr{\bf O}gra{\bf M}ming}, 
which learns to reweight the labeling functions, thereby helping differentiate the noisy LFs from the cleaner and more effective ones. 

The reweighting is achieved by framing the problem in terms of bi-level optimization. 
We argue that using a small \lflabel{} 
can help improve label prediction over hitherto unseen test instances when the \lflabel{} is bootstrapped for (i) inducing LFs, (ii) semi-supervision, and (iii) bi-level optimization to reweight the LFs. 
For most of this work,  the LFs are induced automatically by leveraging part of the approach described in~\cite{varma2018snuba}.
The LFs are induced on the entire \lflabel{}, whereas the semi-supervision and reweighting are performed on the \sslabel{} and \vallabel{} respectively (which are disjoint partitions of \lflabel{}). 

\begin{figure}
    \centering
    \includegraphics[width=0.45\textwidth]{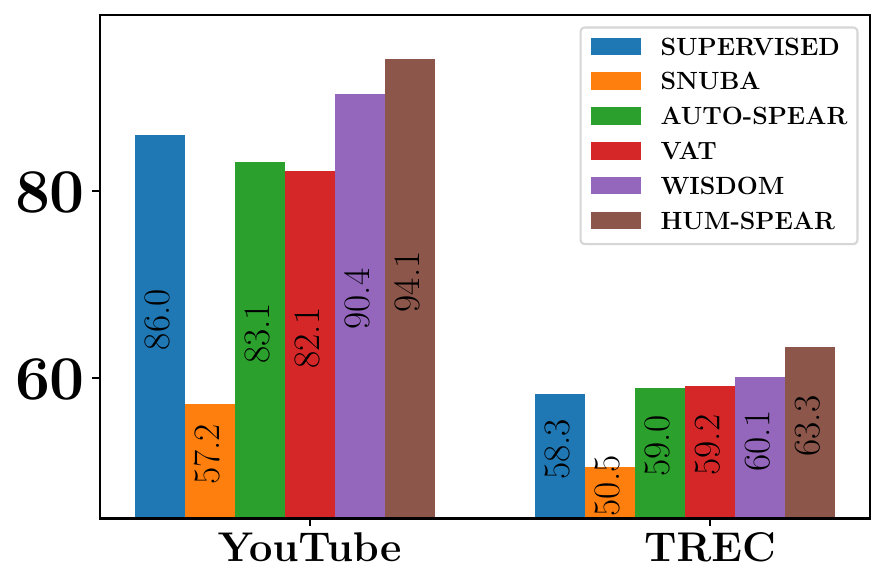}
    \caption{A summary plot contrasting the performance gains obtained using \ouralgo{} on previous state-of-the-art approaches on YouTube and TREC (using Lemma features). \ouralgo{} outperforms other learning approaches with auto-generated LFs. }
    \label{fig:summaryplot}
\end{figure}

\textbf{Our Contributions} are as follows:
 While leveraging \snuba{}~\cite{varma2018snuba} only for \textit{automatically generating} LFs, we address the important limitations of \snuba{} by (i) effectively using the labeled set in a semi-supervised manner using \spear{}~\cite{spear}, and (ii) critically making the labeling function aggregation more \emph{robust} via a reweighting framework. We do the reweighting by using our proposed  bi-level optimization algorithm that weighs each LF separately, giving low importance to noisy LFs and high importance to relevant LFs. We present evaluations on six text classification datasets and show that \ouralgo{} demonstrates better performance than current label aggregation approaches with automatically  (or even human) generated labeling functions. 

A summary of the results are presented in Figure~\ref{fig:summaryplot}. As mentioned, \snuba{} performs worse than a simple supervised baseline that  trained only on the labeled data component. Furthermore, \ouralgo{} outperforms  \textsc{Vat} (a state-of-the-art semi-supervised learning algorithm) and \textsc{Hum-}\spear{} sometimes (a state-of-the-art semi-supervised data programming algorithm with human-generated LFs), demonstrating the benefit of having both semi-supervision and robust LF reweighting with the auto-generated LFs. Finally, \ouralgo{} gets to within 2 - 4\% of \textsc{Hum-Spear} (using human crafted-LFs), without having to incur the cost of generating labeling functions manually, and which can also require significant domain knowledge. 

\section{Background}
\label{background}

\subsection{Notations} \label{sec:notations}
Let us denote the feature space by $\Xcal$ and the label space by $\Ycal \in \{1...K\}$ where $K$ is the number of classes. Let the automatically (or manually) generated labeling functions be denoted by $\lambda_1$ to $\lambda_m$ where $m$ is the number of labeling functions generated. Let the vector $\bfl_{i} = (l_{i1}, l_{i2}, \dots , l_{im})$ denote the firings of all the LFs on an instance $\bfx_i$. Each $l_{ij}$ can be either $1$ or $0$; $l_{ij}=1$ indicates that the LF $\lambda_j$ has fired ({\em i.e.}, triggered) on the instance $x_i$ and $0$ indicates it has not. Furthermore, each labeling function $\lambda_j$ is associated with some class $k_j$ and for an input $x_i$, it outputs 
the label $\tau_{ij} = k_j$
when triggered ({\em i.e.}, $l_{ij} = 1$) 
and $\tau_{ij} = 0$ otherwise. 

Let the \lflabel{} be denoted by $\Lcal = \{(x_i, y_i)\}$ where $i \in \{1 \cdots N \}$ and $N$ is the number of points in \lflabel{}. Similarly, we have an unlabeled dataset denoted as $\Ucal = \{x_i\}$ where $i \in \{N+1 \cdots M\}$ and $M-N$ is the number of unlabeled points. The \lflabel{} is further split into two disjoint sets called \sslabel{} and \vallabel{}. Let the \sslabel{} be denoted by $\Scal = \{(x_i, y_i)\}$ where $i \in \{1 \cdots N/2\}$. Let $\Vcal = \{ (x_i, y_i)\}$ denote the \vallabel{}, where $i \in \{N/2+1 \cdots N\}$. 


\begin{table}[t]
\centering
\begin{adjustbox}{width=0.49\textwidth}
\begin{tabular}{|c|p{10cm}|}
\hline
\textbf{Notation}      & \textbf{Description}                   \\ \hline
$\bfl_{i} \in \{0,1\}^m $ & Firings of all the LFs, $\lambda_1..\lambda_m$ on an instance $\bfx_i$ \\ \hline
$\tau_{ij} \in [0, K]$ & class $k_j$ associated by LF $\lambda_j$, when triggered ($l_{ij} = 1$) on $x_i$ \\ \hline
$f_\phi$ & The feature-based model with parameters $\phi$ operating on feature space $\Xcal$ and on label space $\Ycal \in \{1...K\}$\\ \hline
$P_\theta$               & The label probabilities as per the LF-based aggregation model  with parameters $\theta$                             \\ \hline
\lflabel{} ($\Lcal$) & The entire labeled dataset: $\Lcal = \{(x_i, y_i)\}$ where $i \in \{1 \cdots N \}$. This is used to induce the LFs \\ \hline
\sslabel{} ($\Scal$) & Subset of $\Lcal$ that is used for semi-supervision: $\Scal = \{(x_i, y_i)\}$ where $i \in \{1 \cdots N/2\}$ \\ \hline
\bilabel{} ($\Vcal$) & Subset of $\Lcal$ that is used for reweighting the LFs using a bi-level optimization formulation:  $\Vcal = \{(x_i, y_i)\}$ where $i \in \{N/2+1 \cdots N\}$ \\ \hline
{\emph unlabeled-set} ($\Ucal$) & Unlabeled set: $\Ucal = \{x_i\}$ where $i \in \{N+1 \cdots M\}$ . It is labeled using the induced LFs \\ \hline
$\mathcal{L}_{ce}$ & Cross Entropy Loss  \\ \hline 
$H$                      & Entropy function 
\\ \hline 
$g$                      & Label Prediction from the LF-based graphical model               \\ \hline
$LL_s$                  & Supervised negative log  likelihood over the parameters $\theta$ of the LF aggregation model                     \\ \hline
$LL_u$                  & Unsupervised negative log likelihood summed over labels \\ \hline
KL                     & KL Divergence between two probability models \\ \hline
$R$                      & Quality Guide based loss                               \\ \hline
$\mathcal{L}_{ss}(\theta, \phi, \mathbf{w})$ & The semi-supervised bi-level optimization objective with additional weight parameters $\mathbf{w}$ over the LFs \\ \hline
\end{tabular}
\end{adjustbox}
\caption{Summary of notations used in this paper.}
\label{tab:notations}
\end{table}

\subsection{\snuba: Automatic LF Generation} \label{sec:snuba}
\citet{varma2018snuba} present \snuba{}, a three step approach that (i) automatically generates candidate LFs (referred to as heuristics) using a \lflabel, (ii) filters  heuristics based on  diversity and accuracy metrics to select only relevant heuristics, and (iii) uses the final set of filtered LFs (heuristics) and a label aggregator to compute class probabilities for each point in the unlabeled set $\Ucal$. Steps (i) and (ii) are repeated until the labeled set is exhausted or a limit on the number of iterations is reached. Each LF is a basic composition of propositions on the labeled set. 
A proposition could be a word, a phrase, or a lemma ({\em c.f.}, the second column of Table~\ref{tab:compare-lfs}), or an abstraction such as a part of speech tag. The composition is in the form of a classifier such as a decision stump (1-depth decision tree) or logistic regression.
 
Our \ouralgo{} framework utilizes 
\snuba{} for generating the LFs and thereafter reweigh the LFs via our reweighting framework while jointly learning the model parameters and the LF aggregation in a semi-supervised manner. 

\subsection{\spear: Joint SSL Data Programming} \label{sec:semisup}
\normalsize
\citet{spear} propose a joint learning framework called \spear{} that learns the parameters of a feature-based classification model 
and of the label aggregation model (the LF model) 
in a semi-supervised manner. \spear{} has a feature-based classification model $f_\phi(\bfx)$ that takes the features as input and predicts the class label. \spear{} employs two kinds of models: a logistic regression and a two-layer neural network model. For the LF aggregation model, \spear{} uses an LF-based graphical model inspired from \cage{}~\cite{oishik}. 
\cage{} aggregates the LFs by regularizing parameters such that learned joint distribution of $y$ and $\tau_j$ matches the user provided quality guides over all $y$.
\begin{align}
\vspace{-0.2cm}
{
    P_{\theta}(i, y) = \frac{1}{Z_{\theta}} \prod_{j=1}^{j=m} \psi_\theta(\tau_{ij}, y) }
\end{align}

There are $K$ parameters  $\theta_{j1},\theta_{j2}...\theta_{jK}$ for each LF $\lambda_j$, where $K$ is the number of classes. 
The potential $\psi_\theta$ used in the \cage{} model is defined as:
\begin{align}
\vspace{-0.2cm}
{
\label{dis-pot}
    \psi_\theta(\tau_{ij}, y)= \begin{cases}
    \exp(\theta_{jy}) & \text{if } \tau_{ij}  \neq 0 \\
    1 & \text{otherwise}
    \end{cases}}
\end{align}

The loss function of \spear{} has six terms. These include the cross entropy on the labeled set, an entropy SSL term on the unlabeled dataset, a cross entropy term to ensure consistency between the feature model and the LF model, the LF graphical model terms on the labeled and unlabeled datasets, a KL divergence again for consistency between the two models, and finally a regularizer. The objective function is:
{
\begin{align}
& \sum_{i \in \Lcal} \mathcal{L}_{ce}(f_{\phi}(x_i), y_i) + \sum_{i \in \Ucal} H(f_{\phi}(x_i)) + \nonumber \\
& \sum_{i \in \Ucal} \mathcal{L}_{ce}(f_{\phi}(x_i), g(l_i)) 
 + LL_s(\theta | \Lcal)  + LL_u(\theta | U) + \nonumber \\ & \sum_{i \in \Ucal \cup \Lcal} KL(P_{\theta}(l_i), f_{\phi}(x_i))  + R(\theta | \{q_j\}) 
 \label{eq:spear}
\end{align}
}
where $g$ is the label prediction from the LF-based graphical model. The second component $H()$ models  semi-supervision~\cite{grandvalet2005semi} in the form of  minimization of the entropy of the predictions on the unlabeled dataset $\Ucal$. It provides some semi-supervision by trying to increase the confidence of the predictions made by the model on the unlabeled dataset. (Refer Table \ref{tab:notations} for notations used in the objective function).
In the objective function above, the LF model parameters are $\theta$ while the feature model parameters are $\phi$. The learning problem in \spear{} is simply to optimize the objective jointly over $\theta$ and $\phi$. (We refer readers to ~\citet{spear} for details.)

\noindent \textbf{CAGE loss formulation}: 
The learning problem proposed in \cage{}~\cite{oishik} is a special case of \spear{} where they just use the fifth loss term $LL_u(\theta | U)$ along with the quality guide $R(\theta | \{q_j\})$. The specific loss formulation of \cage{} is as given below:
{
\begin{align}
& LL_u(\theta | U) + R(\theta | \{q_j\})
\label{eq:CAGE}
\end{align}}



\section{The \ouralgo{} Workflow} \label{sec:bilevel} 
In this section, we present our robust aggregation framework for automatically generated LFs. We present the LF generation approach followed by our reweighting algorithm, which solves a bi-level optimization problem. In the bi-level optimization, we learn the LF weights in the outer level, and in the inner level, we learn the feature-based classifier's and labeling function aggregator's parameters jointly. We describe the main components of the \ouralgo{} workflow below (see also Figure~\ref{fig:arch}).  A detailed pseudocode of \ouralgo{} is provided in Algorithm~\ref{alg:wisdom}. We describe the different components of \ouralgo{} below.

\noindent \textbf{Automatic LF Generation using SNUBA}:
Our \ouralgo{} framework utilizes steps (i) and (ii) from \snuba{} ({\em c.f.}, Section~\ref{sec:snuba}) for automatically inducing LFs. That is, it initially iterates between i) candidate LF generation on \lflabel{} $\Lcal$ and ii) filtering them based on diversity and accuracy based criteria, until a limit on the number of iterations is reached (or until the labeled set is completely covered).  We refer to these steps as \snubaLFGen.

\noindent \textbf{Re-Weighting \cage{}}:
To deal with noisy labels effectively, we associate each LF $\lambda_j$ with an additional weight parameter $w_j \in [0, 1]$ that acts as its reliability measure. The $w$'s are optimized on the validation set and have interactions amongst themselves, unlike $\theta$ which is learned on the combination of unlabeled and training sets. 
The discrete potential in \cage{} ({\em c.f.}, eq.\eqref{dis-pot}) can be modified to include weight parameters as follows:
\begin{align}
{
\label{wtdispot}
    \psi_\theta(\tau_{ij}, y)= \begin{cases}
    \exp(w_j\theta_{jy}) & \text{if } \tau_{ij}  \neq 0 \\
    1 & \text{otherwise}
    \end{cases}}
\end{align}

 We observe that if the weight of the $j^{th}$ LF is zero ({\em i.e.}, $w_j=0$), the corresponding weighted potential in eq.~\eqref{wtdispot}
 becomes one, which in turn implies that the $j^{th}$ LF is ignored while maximizing the log-likelihood during label aggregation. Similarly, if all the LFs are associated with a weight value of one ({\em i.e.}, $w_j = 1$), the above weighted potential will degenerate to the discrete potential used in \cage{}. The re-weighted \cage{} is implicitly invoked on lines 12, 13, 17 and 18 of Algorithm~\ref{alg:wisdom} where $\Lcal_{SS}(\theta,\phi,\mathbf{w})$ is invoked.  We compare performance of \cage{} with a bi-level variation in Table \ref{tab:cage-compare}.

\begin{algorithm} [!htbp]
\small{
\DontPrintSemicolon
\KwIn{$\mathcal{L}, \mathcal{S}, \mathcal{V}, \mathcal{U}$, Learning rates: $\alpha, \beta$}
\KwOut{$\theta, \phi, \mathbf{w}$} 
 \SetKwBlock{Begin}{function}{end function}
 {  
   \textcolor{blue}{**** Automatic LF generation using SNUBA **** } \;
   ${\lambda_1, \cdots, \lambda_m}$ = \snubaLFGen($\mathcal{L}$) \;
   Get LFs trigger matrix $\mathbf{l}^s, \mathbf{l}^u$ for sets $\mathcal{S}, \mathcal{U}$ using ${\lambda_1, \cdots, \lambda_m}$\;
   Get LFs output label matrix $\mathbf{\tau}^s, \mathbf{\tau}^u$ for sets $\mathcal{S}, \mathcal{U}$ using ${\lambda_1, \cdots, \lambda_m}$ \;
   \textcolor{blue}{**** The Reweighted Joint SSL **** } \;
   $t=0$; \;
   Randomly initialize model parameters $\theta_{0}$, $\phi_{0}$ and LF weights $\mathbf{w}_{0}$; \;
  \Repeat{convergence}
  {
    Sample mini-batch $s = {(x_i^s, y_i^s, \mathbf{\tau}_i^s, \bfl_i^s)}$, $u = {(x_i^u,  \mathbf{\tau}_i^u, \bfl_i^u)}$ of batch size $B$ from $\{\mathcal{S}, \mathbf{\tau}^s, \bfl^s \}, \{\mathcal{U}, \mathbf{\tau}^u, \bfl^u \}$ \;
    \vspace{1mm}
    \textcolor{blue}{**** Bi-level Optimization **** } \;
    \Indp 
    \textcolor{gray}{**** Inner level **** } \;
    ${\theta}^{*}_{t} = \theta_{t} - \alpha \nabla_{\theta} \mathcal{L}_{ss}(\theta_{t}, \phi_{t}, \mathbf{w}_{t})$ \;
    ${\phi}^{*}_{t} = \phi_{t} - \alpha \nabla_{\phi} \mathcal{L}_{ss}(\theta_{t}, \phi_{t}, \mathbf{w}_{t})$ \;
    \textcolor{gray}{**** Outer level **** } \;
    $\mathbf{w}_{t+1} = \mathbf{w}_{t} - \beta \nabla_{\mathbf{w}} \frac{1}{|\Vcal|} \underset{i \in \Vcal}{\sum}\mathcal{L}_{ce}(f_{{\phi}^{*}_{t}}(x_i), y_i)$ \;
    \vspace{-2.5mm}
    \Indm \textcolor{blue}{**** Update net parameters $\phi, \theta$ ****} \;
    \Indp
    ${\theta}_{t+1} = \theta_{t+1} - \alpha \nabla_{\theta} \mathcal{L}_{ss}(\theta_{t}, \phi_{t}, \mathbf{w}_{t+1})$ \;
    ${\phi}_{t+1} = \phi_{t+1} - \alpha \nabla_{\phi} \mathcal{L}_{ss}(\theta_{t}, \phi_{t}, \mathbf{w}_{t+1})$ \;
    $t = t+1$ \;
  }
 \Return{$\theta_{t+1}, \phi_{t+1}, \mathbf{w}_{t+1}$}
 }
\caption{\ouralgo{}}\label{alg:wisdom}
}
\end{algorithm}

\noindent \textbf{The Reweighted Joint SSL}:
Since the label aggregator graphical model is now dependent on the additional LF weight parameters $\mathbf{w}$, the joint semi-supervised learning objective function is modified as follows:

\small{
\begin{align}
\label{wt-spear}
& \mathcal{L}_{ss}(\theta, \phi, \mathbf{w}) = \sum_{i \in \Scal} \mathcal{L}_{ce}(f_{\phi}(x_i), y_i) + \sum_{i \in \Ucal} H(f_{\phi}(x_i))
\nonumber \\
& +\sum_{i \in \Ucal} \mathcal{L}_{ce}(f_{\phi}(x_i), g(l_i, \mathbf{w}))
 + LL_s(\theta, \mathbf{w} | \Scal) \nonumber \\ 
 & + LL_u(\theta, \mathbf{w} | \Ucal) + \sum_{i \in \Ucal \cup \Scal} KL(P_{\theta, \mathbf{w}}(l_i), f_{\phi}(x_i)) \nonumber \\
 & + R(\theta, \mathbf{w} | \{q_j\}) \hspace{2cm}
\end{align}}
\normalsize

In Section~\ref{sec:appendixexpansion}, we present the somewhat intuitive expansions of terms that are dependent on $\mathbf{w}$.
\noindent \paragraph{Bi-Level Objective:}
\ouralgo{} jointly learns the LF weights and weighted labeling aggregator and feature classifier parameters for the objective function defined in Equation~\eqref{wt-spear}. The LF weights are learned by \ouralgo{} by posing a bi-level optimization problem for this objective function as defined in eq.~\eqref{bilevel-opt} and employing alternating one-step gradient updates. As evident in eq.~\eqref{bilevel-opt}, \ouralgo{} uses a \vallabel{} ($|\Vcal|$) which is a subset of \lflabel{} ($|\Lcal|$) to learn the LF weights. Furthermore, the introduced weight parameters allow filtering of LFs based on the feature model and a bilevel objective in the form of a cross-entropy loss of feature model predictions on the validation set. In essence, \ouralgo{} tries to learn LF weights that result in minimum validation loss on the feature model that is jointly trained with weighted labeling aggregator.

{
\begin{align}
    \label{bilevel-opt}
    &{\mathbf{w}}^{*} = \underset{\mathbf{w}}{\operatorname{argmin}} \frac{1}{|\Vcal|} \sum_{i \in \Vcal} \mathcal{L}_{ce}(f_{{\phi}^{*}}(x_i), y_i) \nonumber \\
    & \text{where\hspace{2mm}} \boldsymbol{\phi}^{*}, \boldsymbol{\theta}^{*} = \underset{\boldsymbol{\phi}, \boldsymbol{\theta}}{\operatorname{argmin}}\ \mathcal{L}_{ss}(\theta, \phi, \mathbf{w})
\end{align}}


However, determining the optimal solution to the above Bi-level objective function is computationally intractable. Hence, inspired by MAML~\cite{finn2017modelagnostic}, \ouralgo{} adopts an iterative alternative minimizing framework, wherein we optimize the objective function at each level using a single gradient descent step. As shown in Algorithm~\ref{alg:wisdom}, lines 12 and 13 are the inner level updates where the parameters $\theta, \phi$ are updated using the current choice of weight parameters $\mathbf{w}$ for one gradient step, and in line 15, the weight parameter $\mathbf{w}$ is updated using the one-step updates from lines 12 and 13. Finally, the net parameters $\phi, \theta$ are updated in lines 17 and 18. This procedure is continued till convergence ({\em e.g.}, no improvement in the outer-level loss) or for a fixed number of epochs. 



\begin{table}[!h]
\centering
\label{tab:dataset}
\begin{adjustbox}{width=0.45\textwidth, height=1.65cm}
\begin{tabular}{lccccc}
\toprule
Dataset & $|\Scal|$ & $|\Vcal|$ & $|\Ucal|$ & \#LFs    & \#Class     \\ 
\midrule
IMDB    & 71 & 71    & 1278  & 18       & 2      \\
YouTube & 55 & 55    & 977   & 11      & 2      \\
SMS     & 463 & 463   & 8335  & 21     &2       \\
TREC    & 273 & 273   & 4918  & 13     &6       \\
Twitter & 707 & 707 & 12019 & 25& 3\\
SST-5 & 568 & 568 & 9651 & 25 & 5 \\
\bottomrule
\end{tabular}
\end{adjustbox}
\caption{Summary statistics of the datasets and the automatically  generated LFs using \snuba{}. 
The test set contains 500 instances for each dataset.} \label{tab:datasetstats}
\end{table} 

\section{Experiments}
\label{sec:expts}
We present evaluations across six datasets 
that we describe in the following Section~\ref{sec:dataset}. In Table~\ref{tab:datasetstats}, we present summary statistics of these datasets, including the sizes of \sslabel{}, \bilabel{} (with \lflabel{} being the union of these disjoint sets) and the number of (auto-generated) LFs used in the experiments.

\subsection{Datasets}
\label{sec:dataset} 
We use the following datasets in our experiments:
(1) \textbf{TREC} \cite{trec}: A question classification dataset with six categories:  \texttt{Description, Entity, Human, Abbreviation, Numeric, Location}. (2) \textbf{YouTube Spam Classification} \cite{youtube}: A spam classification task over comments on YouTube videos.
(3) \textbf{IMDB Genre Classification\footnote{\url{www.imdb.com/datasets}}}: A plot summary based movie genre binary classification dataset.
(4) \textbf{SMS Spam Classification}~\cite{sms}: A binary spam classification dataset to detect spam in SMS messages. (5) \textbf{Twitter Sentiment}~\cite{twitter}: This is a 3-class sentiment classification problem extracted from Twitter feed of popular airline handles. Each tweet is either labeled as negative, neutral, and positive labels. (6) \textbf{Stanford Sentiment Treebank (SST-5)}~\cite{sst5} is a single sentence movie review dataset, with each sentence labeled as either negative, somewhat negative, neutral, somewhat positive, or positive.

\subsection{Baselines}
In Table \ref{tab:main}, we compare our approach against the following baselines:\\
\textbf{Snuba} \cite{snuba}: Recall from Section~\ref{sec:snuba} that \textsc{Snuba} iteratively induces LFs from the count-based raw features of the dataset in the steps (i) and (ii). 
For the step (iii), as in~\cite{snuba}, we employ a generative model to assign probabilistic labels to the unlabeled set. 
These probabilistic labels are obtained by training a 2-layered NN model.\\
\textbf{Supervised (SUP)}: This is the model obtained by training the classifier $P_\theta(y|x)$ only on \lflabel. This baseline does not use the unlabeled set.\\ 
\textbf{Learning to Reweight (L2R)}~\cite{ren2018learning}: This method trains the classifier using a meta-learning algorithm over the noisy labels in the unlabeled set obtained using the automatically generated labeling functions and aggregated using \textsc{Snorkel}. It uses an  online algorithm that assigns importance to examples based on the gradient. \\
\textbf{Posterior Regularization (PR)}~\cite{Hu2016HarnessingDN}: This is a method for joint learning of a rule and feature network in a teacher-student setup. Similarly to L2R, it uses the noisy labels in the unlabeled set obtained using the automatically generated labeling functions. \\
\textbf{Imply Loss (IL)}~\cite{awasthi2020learning}: This method leverages both rules and labeled data by associating each rule with exemplars of correct firings ({\em i.e.}, instantiations) of that rule. Their joint training algorithms de-noise over-generalized rules and train a classification model. This is also run on the automatically generated LFs.\\
\textbf{SPEAR}~\cite{spear}: This method employs a semi-supervised framework combined with a graphical model for consensus amongst the LFs to train the model. We compare against two versions of \spear{}. The first that (just like L2R, PR, IL, and VAT) uses auto-generated LFs (which we call \textsc{Auto-}\spear{}), and the second, {\em viz.}, \textsc{Hum-}\spear{}, which uses the human LFs. \\
\textbf{VAT}: Virtual Adversarial Training~\cite{miyato2018virtual} is a semi-supervised approach that uses the virtual adversarial loss on the unlabeled points, thereby ensuring robustness of the conditional label distribution on the unlabeled points. 

\subsection{Experimental Setting}

To train our model on the \sslabel{}, we use a neural network architecture with two hidden layers (512 units) and ReLU activation function as our feature-based model $f_{\phi}$. We choose our classification network to be the same as \spear{}~\cite{spear}. We consider two types of features: a) raw words and b) lemmatizations, as an input to our supervised model (lemmatization is a technique to reduce a word, \textit{e.g.,} `walking,' into its root form, 'walk'). Additionally, these features are used as basic propositions over which composite LFs are built.

Each experimental run involves training \ouralgo{} for 100 epochs with early stopping based on \vallabel. Our model is optimized using mini-batch gradient descent with the Adam optimizer. We tuned the hyperparameters on the \vallabel{}, and the optimal configuration was found to have a dropout probability of 0.80 and a batch size of 32. Further, the optimal configuration learning rates for the classifier and LF aggregation models were 0.0003 and 0.01, respectively. 
Performance numbers for each experiment are obtained by averaging over five independent runs, each having a different random initialization. For evaluation on the test set, the model with the best performance on the \vallabel{} was chosen. On all datasets, macro-F1 is employed as the evaluation criterion. We implement all our models in PyTorch\footnote{\url{https://pytorch.org/}}~\cite{paszke2019pytorch}. 
We run all our experiments on Nvidia RTX 2080 Ti GPUs with 12 GB RAM set within Intel Xeon Gold 5120 CPU having 56 cores and 256 GB RAM. Model training times range from 15 mins (YouTube) to 100 mins (TREC).

\begin{table*}
\centering
\begin{adjustbox}{width=\textwidth}
    
\begin{tabular}{ccc||ccccccc|c}
\toprule
                         & \multicolumn{10}{c}{Methods}                                                                          \\ 
\cline{3-11}
Dataset                 &       & \ssalgo{} & \snuba{} & L2R    & VAT   & PR & IL & \textsc{Auto-}\spear{} & \ouralgo{} & \textsc{Hum-}\spear{} \\ 
\hline
\multirow{2}{*}{IMDB}    & Raw   & 68.8 {\scriptsize(0.2)} & -5.9 {\scriptsize(2)} &-6.6 {\scriptsize(1.6)} &-12.3  {\scriptsize (1)}  & +2.7 {\scriptsize(15.6)}        & +2.4 {\scriptsize(1.7)}     & +2.4 {\scriptsize(1.6)} &\textbf{ +3.4}  {\scriptsize(0.1)} & {NA}  \\
& Lemma & 72.4 {\scriptsize(1.3)}  & -14.4 {\scriptsize(5.7)}&  -3.7 {\scriptsize(14.7)} &-19.3  {\scriptsize(0.1)} & -11.7 {\scriptsize(4.1)}         & -6.4 {\scriptsize(8.2)}      & -2.4 {\scriptsize(1.6)}  & \textbf{+3.6}  {\scriptsize(1.4)}  & {NA}  \\
\hline
\multirow{2}{*}{YouTube} & Raw   & 90.8 {\scriptsize(0.3)}  & -33.2 {\scriptsize(1.8)} & +0.5 {\scriptsize(0.5)}  & +0.5 {\scriptsize (0)}   & -4.7 {\scriptsize(0.4)}          & +0.2 {\scriptsize(0.3)}      & +0.8 {\scriptsize(0.5)}  & \textbf{+1.4} {\scriptsize (0.0)}  & \color{blue}{+3.8} {\scriptsize(0.2)} \\
                         & Lemma & 86  {\scriptsize(0.3)} &-28.7 {\scriptsize(2.9)}& -2.2 {\scriptsize(0.7)}  &-3.8 {\scriptsize(0.2)} & -7.5 {\scriptsize(0.5)}          & -2.6 {\scriptsize(0.3)}      & -7.9 {\scriptsize(3.7)}  &\textbf{+4.4}  {\scriptsize(0.2)} & \color{blue}{+6.9}{\scriptsize(0.7)} \\
\hline
\multirow{2}{*}{SMS}     & Raw   & 92.3 {\scriptsize(0.5)}  & -16.7 {\scriptsize(9.8)}& -5.6 {\scriptsize(0.4)}  & +1.1 {\scriptsize(0.1)} & +0.3 {\scriptsize(0.1)}          & 0 {\scriptsize(0.3)}        & 0.4 {\scriptsize(0.8)}  & \textbf{+1.5}  {\scriptsize(0.1)}  & \color{blue}{+0.1} {\scriptsize(0.5)}\\
                         & Lemma & 91.4 {\scriptsize(0.5)}  & -16.1 {\scriptsize(5.3)}& -5.9 {\scriptsize(0.5)}  & +1.6 {\scriptsize(0.5)} & +0.6 {\scriptsize(0.3)}          & +1.5 {\scriptsize(0.3)}      & -1.5 {\scriptsize(1.8)}  & \textbf{+2}   {\scriptsize(0.5)}  & \color{blue}{0 }{\scriptsize(0.1)}  \\
\hline
\multirow{2}{*}{TREC}    & Raw   & 58.3 {\scriptsize(3.1)} &-6.8 {\scriptsize(4.1)}& -11.8 {\scriptsize(0.8)} & \textbf{+3.7} {\scriptsize(0.5)} & -2.2 {\scriptsize(0.6)}          & -0.3 {\scriptsize(0.8)}      & -0.9 {\scriptsize(0.5)}  & +3.4   {\scriptsize(0.5)} & \color{blue}{+5} {\scriptsize(0.5)} \\
                         & Lemma & 56.3 {\scriptsize(0.3)} &-5.8 {\scriptsize(5.1)} & -5.5 {\scriptsize(0.6)}  & +3.0 {\scriptsize(0.5)} & +0.4 {\scriptsize(0.4)}          & +0.8 {\scriptsize(0.8)}      & +2.7 {\scriptsize(0.1)}  & \textbf{+3.9}  {\scriptsize(0.5)} & \color{blue}{+4.7}{\scriptsize(0.3)}\\
\hline

\multirow{2}{*}{Twitter}    & Raw   & 52.61 {\scriptsize(0.12)} & -7 {\scriptsize(4.1)}& -5 {\scriptsize(2.3)} & +0.41 {\scriptsize(3.5)} & -4.49 {\scriptsize(3.6)}          & -0.85 {\scriptsize(0.6)}      & -4.24 {\scriptsize(0.4)}  & \textbf{+1.04}   {\scriptsize(0.8)} & {NA} \\
                         & Lemma & 61.24 {\scriptsize(0.52)} & -9.28 {\scriptsize(5.1)} & -18.03 {\scriptsize(1.5)}  & -10.8 {\scriptsize(5.3)} & -8.12 {\scriptsize(2.1)}  & -3.79 {\scriptsize(0.1)}  & +1.9 {\scriptsize(0.1)}  & \textbf{+3.97}  {\scriptsize(0.7)} & {NA}\\
\hline
\multirow{2}{*}{SST-5}    & Raw   & 27.54 {\scriptsize(0.12)} & -9 {\scriptsize(2.2)}& -7.98 {\scriptsize(0.2)} & -6.12 {\scriptsize(0.12)} & -5.59 {\scriptsize(0.2)} & -2.11 {\scriptsize(0.1)}      & -4.12 {\scriptsize(0.1)}  & \textbf{+0.97}   {\scriptsize(0.3)} & {NA} \\
                         & Lemma & 27.52 {\scriptsize(0.52)} &-8.31 {\scriptsize(3.1)} & -8.1 {\scriptsize(8.1)}  & -7.89 {\scriptsize(1.6)} & -7 {\scriptsize(4.7)} & -3.4 {\scriptsize(0.16)}      & -3.13 {\scriptsize(2.1)}  & \textbf{+0.79}  {\scriptsize(0.3)} & {NA}\\
\bottomrule
\end{tabular}
\end{adjustbox}
\caption{Performance of different approaches on six datasets, {\em viz.}, IMDB, YouTube, SMS, TREC, Twitter, and SST-5. Results are shown for both 'Raw' or 'Lemmatized' features. The numbers reported are macro-F1 scores over the test set averaged over 5 runs, and for all methods after the double-line are reported as gains over the baseline (\ssalgo). L2R, PR, IL, \textsc{Auto-}\spear{}, and \ouralgo{} all use the automatically generated LFs; \ssalgo{} and \textsc{Vat} do not use LFs; and \textsc{Hum-}\spear{} uses the {\color{blue} human generated LFs}. 'NA' in \textsc{Hum-}\spear{} column is when human LFs are not available.
Numbers in brackets `()' represents standard deviation of the original scores and not of the gains.}
\label{tab:main}
\end{table*} 

\subsection{Results}
\begin{figure}
    \centering
    \includegraphics[width=0.45\textwidth, height=3.5cm]{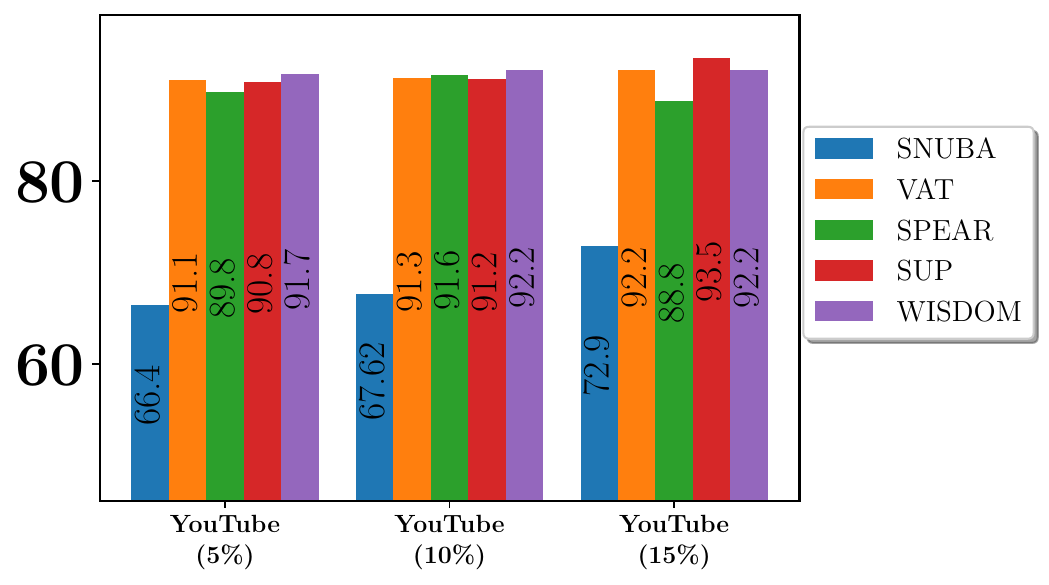}
    \caption{Ablation study with different \lflabel{} sizes on the YouTube dataset.}
    \label{fig:ablation}
\end{figure}

In Table~\ref{tab:main}, we compare the performance of \ouralgo{} against different baselines (all using auto-generated labeling functions except VAT), for both raw and lemmatized count features ({\em c.f.} Section~\ref{sec:snuba}) across multiple datasets. We observe that \snuba{} performs worse than the  \ssalgo{} baseline  on all datasets, exhibiting high variance over different runs (surprisingly, \citet{snuba} did not compare the performance of \snuba{} against the  \emph{supervised} baseline). Learning to Reweight (L2R) performs worse than \ssalgo{} on all datasets except YouTube. Posterior regularization, imply loss and \spear{} show gains over \ssalgo{} on a few datasets, but not consistently across all datasets and settings. Finally, \textsc{Vat} obtains competitive results in some settings ({\em e.g.}, TREC dataset) but performs much worse on others ({\em e.g.}, IMDB and SST-5). In contrast, \ouralgo{} achieves consistent gains over \ssalgo{} and the other baselines in almost all datasets (except TREC with raw features where \textsc{Vat} does slightly better than \ouralgo). Additionally, \ouralgo{} yields smaller variance over different runs compared to other semi-supervised approaches. Recall that the main difference between \ouralgo{} and Auto-\spear{} is that the former reweighs the LFs in both the label aggregator as well as in the semi-supervised loss, as against Auto-\spear{} which does not reweigh the LFs at all. Consequently, the aforementioned empirical gains illustrate the robustness of the bi-level optimisation algorithm. Note that these numbers are all reported using only 10\% labeled data, and hence, results for some datasets (starting with \ssalgo{}) might appear lower than those reported in the literature.
Note that, we compare \ouralgo{} (using automatically induced LFs) against the \textsc{Hum-}\spear{} which uses \textit{the human crafted} LFs in conjunction with the state-of-the-art \spear{} approach~\cite{spear}. 
Although \ouralgo{} uses auto-generated LFs, it sometimes performs better than \textsc{Hum-}\spear{}, which utilizes human-curated LFs. On careful analysis (presented in Section~\ref{sec:appendixhumanVsAuto} of the supplementary), we observe that the human curated LFs tend to be more generic abstractions of possible patterns 
without assessing 
how precise they are for the end task. Consequently, these abstract human-LFs tend to have not only higher collective coverage but also high mutual conflicts and lower average individual precision values than the automatically induced LFs. Given the individual strengths of both {\emph human-lfs} and {\emph auto-lfs}, it might be interesting to consider using them in conjunction with each other in order to improve performance as future work. 
An ablation test in Figure~\ref{fig:ablation} reveals that \ouralgo{} performs well even for small-sized \lflabel{} unlike other baselines, demonstrating its robustness in scenarios with only few labeled examples.


\subsection{Importance of the Bi-Level formulation}
A label aggregation approach, such as \cage{}, \snorkel{}, may improve the consensus labeling across LFs, but not necessarily their agreement with the ground truth. Further, when LFs are noisy (or induced automatically), the performance of the \cage{} model can suffer. 
However, the bi-level framework of \cage{} can alleviate these problems since it implicitly reduces the noise in LFs. In order to demonstrate effectiveness of the bi-level formulation, we compare \cage{}(Eq~\eqref{eq:CAGE}) with two variants (i) \cage{}\textsubscript{val}\footnote{\cage{}\textsubscript{val} - equivalent  to using only $LL_s(\theta | \Lcal) + LL_u\left(\theta, {\mathbf w} | {\mathcal U}\right) + R\left(\theta, {\mathbf w} | \{q_j\}\right)$ in Eq~\eqref{eq:spear}} that considers validation set feedback in the loss formulation for promoting LF agreement with ground-truth label and (ii) \cage{}\textsubscript{Bi-level} with the proposed bi-level formulation that tries to do the same\footnote{In other words, \cage{}\textsubscript{Bi-level} is equivalent to using only $LL_u\left(\theta, {\mathbf w} | {\mathcal U}\right) + R\left(\theta, {\mathbf w} | \{q_j\}\right)$ in Eq~\eqref{wt-spear}}. We present our results in Table~\ref{tab:cage-compare}. The performance of our \cage{}\textsubscript{Bi-level} is clearly superior to the original \cage{} model, as well as to the \cage{}\textsubscript{val} model. Thus, the bi-level formulation more effectively incorporates validation set feedback than other formulations as demonstrated by application of bi-level on both \spear{} as well as on \cage. In Table~\ref{tab:compare-lfs}, we had presented some illustrative examples (from the TREC dataset) of automatically induced LFs whose weights are relatively higher based on the bi-level formulation along with those that are down-weighted owing to their  conflicting signals. We present additional examples as well as further qualitative analysis in Section~\ref{sec:appendixLFQual} of the supplementary.
\begin{table}[]
\centering
\begin{adjustbox}{width=0.38\textwidth, height=1cm}
\begin{tabular}{|l|c|c|c|}
\hline
& Youtube & SMS & TREC\\ \hline
\cage{} & 62.45 & 18.1 & 14.1\\ \hline
\cage{}\textsubscript{val} & 84.62  & 39.61  & 37.99  \\ \hline
\cage{}\textsubscript{Bi-level} & 87.11 & 43.22 & 39.34 \\ \hline
\end{tabular}
\end{adjustbox}
\caption{Comparison of \cage{} model with two variants. \cage{}\textsubscript{val} includes validation set feedback in the original \cage{} loss function and \cage{}\textsubscript{Bi-level} is bi-level formulation of \cage{} objective using Eq \ref{wtdispot}.}
\label{tab:cage-compare}
\end{table}

\section{Related Work} 
In this section, we describe some additional related work that was not covered in Section~\ref{background}.

\noindent \textbf{Automatic Rule Generation}: The programming by examples paradigm produces a program from a given set of input-output pairs~\cite{gulwani2012synthesis, singh2012synthesizing}. It synthesises those programs that satisfy all input-output pairs. RuleNN~\cite{rulenn} learns interpretable first-order logic rules as composition of semantic role attributes. Many of these approaches, however, learn more involved rules (using {\em e.g.}, a neural network) which may not work in the realistic setting of very small labeled data. In contrast, \snuba{} and \ouralgo{}{} use more interpretable models~\cite{Rudin2019Stop} like logistic regression and decision trees for rule induction.

\noindent \textbf{Semi-supervised Learning (SSL)}: The goal of SSL is to effectively use unlabeled data while training. Early SSL algorithms used regularization-based approaches like margin regularization, and laplacian regularization \cite{chapelle}. 
Most recent SSL approaches like Mean Teacher \cite{tarvainen2018mean}, VAT \cite{miyato2018virtual}, UDA \cite{xie2020unsupervised}, MixMatch \cite{berthelot2019mixmatch} and FixMatch \cite{sohn2020fixmatch} introduced various kinds of perturbations and augmentations that can be used along with consistency loss. Even though the current SSL approaches perform well even with minimal labels, they are computationally intensive and cannot be easily implemented in low-resource scenarios. Furthermore, it is tough to explain the discriminative behavior of the semi-supervised models.

\noindent \textbf{Bi-level Optimization}: 
The concept of bi-level optimization has been discussed in  ~\cite{von1952theory, brackenbilevel, bilevel}.
Since then, the framework of bi-level optimization has been used in various machine learning applications like hyperparameter tuning ~\cite{mackay2018selftuning, franceschi2018bilevel, sinha2020gradientbased}, robust learning ~\cite{ren2018learning, pmlr-v119-guo20i}, meta-learning ~\cite{finn2017modelagnostic}, efficient learning ~\cite{killamsetty2020glister} and continual learning ~\cite{borsos2020coresets}. Previous applications of the bi-level optimization framework for robust learning have been limited to supervised and semi-supervised learning settings. To the best of our knowledge, \ouralgo{}  is the first framework that uses a bi-level optimization approach for robust aggregation of labeling functions.

\section{Conclusion}
 While induction of labeling functions (LFs) for data-programming has been attempted in the past by~\citet{snuba}, we observe in our experiments that the resulting model in itself does not perform well on text classification tasks, and turns out to be even worse than the supervised baseline. A more recent semi-supervised data programming approach called SPEAR~\cite{spear}, when used in conjunction with the induced LFs, performs better,
 though it fails to consistently outperform the supervised baseline. 
 In this paper, we introduce 
 \ouralgo, a bi-level optimization formulation for reweighting 
 the LFs, which injects  
 robustness into the semi-supervised data programming approach, thus allowing it to perform well in the presence of noisy LFs. 
 On a reasonably wide variety of text classification datasets, we show that \ouralgo{} consistently outperforms all other approaches, while also coming close to the skyline of \spear{} using human-generated LFs.  

\section*{Acknowledgements and Disclosure of Funding}
We thank anonymous reviewers for providing constructive feedback. Ayush Maheshwari is supported by a Fellowship from Ekal Foundation (www.ekal.org). We are also grateful to IBM Research, India (specifically the IBM AI Horizon Networks - IIT Bombay initiative) for their support and sponsorship. Rishabh Iyer and Krishnateja Killamsetty were funded by the National Science Foundation(NSF) under Grant Number 2106937, a startup grant from UT Dallas, and a Google and Adobe research award. Any opinions, findings, and conclusions or recommendations expressed in this material are those of the author(s) and do not necessarily reflect the views of the National Science Foundation, Google or Adobe.
\bibliography{refs}
\bibliographystyle{acl_natbib}
\onecolumn

\begin{center}
 \Large{\bf Appendix}
\end{center}

\section{Explanation of loss terms}
\label{sec:appendixexpansion}

\noindent \textbf{First Component (L1): } The first component (L1) of the loss $L_{CE}\left(P_\phi^f(y|\bfx_i), y_i\right) = -\log\left(P_\phi^f(y=y_i|\bfx_i)\right)$ is the standard cross-entropy loss on the labelled dataset $\Lcal$ for the model $P_\phi^f$.  

\noindent \textbf{Second Component (L2): } The second component L2 is the semi-supervised loss on the unlabelled data $\Ucal$. In our framework, we can use any unsupervised loss function. 

\noindent \textbf{Third Component (L3): } The third component $L_{CE}\left(P_\phi^f(y|\bfx_i), g(\bfl_i), w\right)$ is the cross entropy of the classification model using the hypothesised labels from CAGE ~\cite{oishik} on $\Ucal$. 
Given that $\bfl_i$ is the output vector of all labelling functions for any $\bfx_i \in \Ucal$, we specify the predicted label for $\bfx_i$ using the LF-based graphical model $P_\theta(\bfl_i, y)$  as:
    $g(\bfl_i) = \underset{y}{\mbox{argmax}} P_{\theta,w}(\bfl_i, y)$ \\
\noindent \textbf{Fourth Component (L4): } The fourth component $LL_s(\theta|\Lcal)$ is the (supervised) negative log likelihood loss on the labelled dataset $\Lcal$:  
    $LL_s(\theta, w|\Lcal) = - \sum \limits _{i=1}^{N} \log P_{\theta, w}(\bfl_i, y_i)$

\noindent \textbf{Fifth Component (L5): } The fifth component $LL_u(\theta, w|\Ucal)$ is the negative log likelihood loss for the unlabelled dataset $\Ucal$. Since the true label information is not  available, the probabilities need to be summed over $y$:  
    $LL_u(\theta, w|\Ucal) = - \sum \limits _{i=N+1}^{M} \log \sum \limits _{y \in \Ycal} P_{\theta, w}(\bfl_i, y)$

\noindent \textbf{Sixth Component (L6): } The sixth component $KL(P_{\phi,w}^f(y|\bfx_i), P_\theta(y|\bfl_i))$ 
is the Kullback-Leibler (KL) divergence between the predictions of both the models, {\em viz.}, feature-based model $f_\phi$ and the LF-based graphical model $P_\theta$  summed over every example $\bfx_i \in \Ucal \cup \Lcal$. Through this term, we try and make the models agree in their predictions over the union of the labelled and unlabelled datasets.

\textbf{Quality Guides (QG): } As a last component in our objective, 
we use quality guides  $R(\theta, w|\{q_j\})$ on LFs which have been shown~\cite{oishik}  to stabilise the unsupervised likelihood training while using labelling functions. Let $q_j$ be  the fraction of cases where $\lambda_j$  correctly triggered. And let $q_j^t$ be the user's belief on the fraction of examples $\bfx_i$ where $y_i$ and $l_{ij}$ agree. If user's beliefs weren't available, we consider precision of LFs on validation set as the user's beliefs. Except SMS dataset, we take precision of LFs on validations set as quality guides. If $P_{\theta, w}(y_i=k_j|l_{ij}=1)$ is the model-based precision over the LFs, the quality guide based loss  can be expressed as 
$R(\theta, w | \{q_j^t\}) = - \bigg(\sum_j  q_j^t \log P_{\theta, w}(y_i=k_j|l_{ij}=1) + (1-q_j^t)  \log (1-P_{\theta, w}(y_i=k_j|l_{ij}=1))\bigg)$.

\section{LF Analysis}
\label{sec:appendixhumanVsAuto}

We compare statistics of automatically induced LFs and human-curated LFs in Table~\ref{tab:appendix-lf-analysis}. While developing LFs, humans generally tend to design LFs based on generalizibility of the pattern without worrying much about the conflicts among the patterns. Whereas the LF induction in \ouralgo{}
focuses on inducing individually precise LFs without necessarily focusing on the overall coverage.
Except in the case of the SMS dataset, collective coverage of human designed LFs is much higher than that of the automatically induced LFs. We also observe in Table~\ref{tab:appendix-lf-analysis} that higher coverage leads to higher conflicts. Whereas, on an average, the precision is higher for each of the  automatically induced LFs in the case of  every dataset. 

\begin{table}[!h]
\centering
\begin{tabular}{|l|cccc|cccc|}
\hline
\textbf{} & \multicolumn{4}{c|}{\textbf{Auto LFs}} & \multicolumn{4}{c|}{\textbf{Human LFs}} \\ \hline
\textbf{} & \multicolumn{1}{c|}{\textbf{\#LFs}} & \multicolumn{1}{c|}{\textbf{Precision}} & \multicolumn{1}{c|}{\textbf{Conflict}} & \textbf{Cover (\%)} & \multicolumn{1}{c|}{\textbf{\#LFs}} & \multicolumn{1}{c|}{\textbf{Precision}} & \multicolumn{1}{c|}{\textbf{Conflict}} & \textbf{Cover(\%)} \\ \hline
YouTube & \multicolumn{1}{c|}{11} & \multicolumn{1}{c|}{94.3} & \multicolumn{1}{c|}{8.1} & 63.4 & \multicolumn{1}{c|}{10} & \multicolumn{1}{c|}{79.8} & \multicolumn{1}{c|}{28.7} & 88.0 \\ \hline
SMS & \multicolumn{1}{c|}{25} & \multicolumn{1}{c|}{94.9} & \multicolumn{1}{c|}{3.2} & 47.9 & \multicolumn{1}{c|}{73} & \multicolumn{1}{c|}{92.3} & \multicolumn{1}{c|}{1.0} & 33.3 \\ \hline
TREC & \multicolumn{1}{c|}{13} & \multicolumn{1}{c|}{70.1} & \multicolumn{1}{c|}{2.3} & 62.3 & \multicolumn{1}{c|}{68} & \multicolumn{1}{c|}{59.9} & \multicolumn{1}{c|}{22.3} & 95.1 \\ \hline
\end{tabular}
\caption{Comparison of automatically generated LFs with human-curated LFs. Coverage is fraction of instances in $\mathcal{U}$ covered by at least one rule. Precision refers to micro precision of rules. Conflict denotes the fraction of instances covered by conflicting rules among all the covered instances. }
\label{tab:appendix-lf-analysis}
\end{table}

\section{Qualitative Analysis of Automatically Induced LFs}
\label{sec:appendixLFQual}
For the six datasets used for experimentation, we automatically induce LFs using Snuba \cite{snuba}. We show the automatically induced LFs and their respective weights assigned by \ouralgo{} for three datasets TREC, IMDB, and SMS below.

\begin{table}[!h]
\centering
\begin{tabular}{|l|l|l|}
\hline
\multicolumn{1}{|c|}{Class} & \multicolumn{1}{c|}{LF} & \multicolumn{1}{c|}{Weights} \\ 
\hline
NUM & how many & 1 \\ \hline
NUM & how & 1 \\ \hline
NUM & many & 0.62\\ \hline\hline
DESC & what kind & 1\\ \hline
DESC & what was & 0.54\\ \hline \hline
LOC & city & 1\\ \hline
LOC & country & 0.84 \\ \hline
LOC & where & 0.05\\ \hline \hline
ENTY & what does & 1\\ \hline
ENTY & def & 1 \\ \hline
ENTY & why & 0.8\\ \hline
ENTY & what is & 0.65\\ \hline \hline
HUM & who & 0.00012\\ \hline
\hline
\end{tabular}%
\caption{Automatically induced LFs by Snuba \cite{snuba} for the  TREC dataset sorted in descending order of weights per class assigned by \ouralgo{}. Column 1 refers to the class associated with the induced LF.  No LFs were induced for class \texttt{Abbreviation}.}
\label{tab:lfs-trec}
\end{table}

In Table \ref{tab:lfs-trec}, we present LFs produced by the Snuba for the TREC dataset sorted in descending order of weights for each class along with the weights assigned by \ouralgo{} to each of the LFs. From analysis, we observe that \ouralgo{} does a good job of reweighting LFs. For instance, \texttt{how many} was given higher weightage than \texttt{how} and \texttt{many} for class Numeric; this sounds logical as well since sentences containing the keyword \texttt{how many} are more likely to belong to class Numeric than sentences containing the keyword \texttt{how} or \texttt{many}. Another example is among LFs associated with Location class, LFs \texttt{city} and \texttt{country} were given higher weightage than \texttt{where}. However, \ouralgo{} does a poor job by assigning a very small weight value to the single LF \texttt{who} associated with the Human class. 

\begin{table}[]
\centering
\begin{tabular}{|l|l|l|}
\hline
\multicolumn{1}{|c|}{Class} & \multicolumn{1}{c|}{LF} & \multicolumn{1}{c|}{Weights} \\ 
\hline
ROMANCE & wife & 0.412\\ \hline
ROMANCE & love & 0.042\\ \hline
ROMANCE & boyfriend & 0  \\ \hline
ROMANCE & friendship & 0 \\ \hline
ROMANCE & wealthy & 0\\ \hline
ROMANCE & story & 0 \\ \hline
ROMANCE & town & 0\\ \hline
ROMANCE & friend & 0\\ \hline
\hline
ACTION & government & 1 \\ \hline
ACTION & plan & 0.985 \\ \hline
ACTION & agent & 0.913 \\ \hline
ACTION & team & 0.753\\ \hline
ACTION & race & 0.685\\ \hline
\hline
\end{tabular}
\caption{Automatically induced LFs by Snuba \cite{snuba} for the IMDB dataset sorted in descending order of weights per class assigned by \ouralgo{}. Column 1 refers to the class associated with the induced LF.}
\label{tab:lfs-imdb}
\end{table}

In Table \ref{tab:lfs-imdb}, we present LFs produced by the Snuba for the IMDB dataset sorted in descending order of weights for each class along with the weights assigned by \ouralgo{} to each of the LFs. For the IMDB dataset as well, we can see that \ouralgo{} does a good job of reweighting LFs. For instance,among the LFs associated with the class ROMANCE, \texttt{wife} and \texttt{love} were given higher weightage than other LFS like \texttt{friendship}, \texttt{wealthy}, \texttt{town}; this sounds logical as well since ROMANCE is often associated with the sentences containing the keywords \texttt{wife}, \texttt{love} than sentences containing the keyword \texttt{friendship}, \texttt{town}, \texttt{wealthy}. One more key observation is that apart from LFs \texttt{wife} and \texttt{love}, all other LFs associated with the class ROMANCE are given weights of $0$(equivalent to ignoring them). However, assigning $0$ weights is controversial for LFs like \texttt{boyfriend} since there is a possibility of ROMANCE associated with the sentence containing keyword \texttt{boyfriend}. Similarly for LFs associated with Action class, LFs  \texttt{government}, \texttt{agent}, and \texttt{plan} were given higher weightage than \texttt{race}, and \texttt{team}.

\begin{table}[]
\centering
\begin{tabular}{|l|l|l|}
\hline
\multicolumn{1}{|c|}{Class} & \multicolumn{1}{c|}{LF} & \multicolumn{1}{c|}{Weights} \\ 
\hline
SPAM & ur & 1\\ \hline
SPAM & video & 1 \\ \hline
SPAM & com & 1 \\ \hline
SPAM & contact & 0.2213 \\ \hline
SPAM & holiday & 0.1593\\ \hline
SPAM & free & 0 \\ \hline
SPAM & claim & 0 \\ \hline
SPAM & stop & 0 \\ \hline
SPAM & won & 0 \\ \hline
SPAM & win & 0 \\ \hline
SPAM & uk & 0\\ \hline
SPAM & text & 0\\ \hline
SPAM & urgent & 0\\ \hline
NOTSPAM & come & 1 \\ \hline
NOTSPAM & ok & 1\\ \hline
NOTSPAM & got & 1 \\ \hline
NOTSPAM & like & 1 \\ \hline
NOTSPAM & sorry & 0.03731254\\ \hline
\end{tabular}
\caption{Automatically induced LFs by Snuba \cite{snuba} for the SMS dataset sorted in descending order of weights per class assigned by \ouralgo{}. Column 1 refers to the class associated with the induced LF.}
\label{tab:lfs-sms}
\end{table}

In Table \ref{tab:lfs-sms}, we present LFs produced by the Snuba for the SMS dataset sorted in descending order of weights for each class along with the weights assigned by \ouralgo{} to each of the LFs. For the SMS dataset, we can see that \ouralgo{} did not do as good a job of reweighting as done on other datasets. For instance,among the LFs associated with the class SPAM, \texttt{ur}, \texttt{video} and \texttt{cam} were given higher weightage while completely ignoring(i.e., assigned a weight of zero) to other important LFS like \texttt{free}, \texttt{claim}, \texttt{won}. Whereas for LFs associated with the class NOT SPAM, \ouralgo{} did a good job. One possible reason for the poor job of \ouralgo{} for reweighting LFs associated with the class SPAM is that class imbalance present in the unlabeled set, where the sample count of samples of the class SPAM is eight times smaller than the sample count of the class SPAM. From our LF analysis results across the three datasets, we observe that \ouralgo{} tries to up weigh LFs that are more specific and precise and downweigh LFs that are abstract and less precise.

\end{document}